\documentclass[11pt]{article}
\PassOptionsToPackage{hyphens}{url}\usepackage[hyperref]{acl2019}
\usepackage{times}
\aclfinalcopy
\usepackage{latexsym}
\usepackage{xcolor}
\usepackage{graphicx}
\usepackage{amsmath}
\usepackage{siunitx}
\usepackage{booktabs}
\usepackage{multirow}

\usepackage{arydshln}
\usepackage{enumitem}

\makeatletter
\def\adl@drawiv#1#2#3{%
        \hskip.5\tabcolsep
        \xleaders#3{#2.5\@tempdimb #1{1}#2.5\@tempdimb}%
                #2\z@ plus1fil minus1fil\relax
        \hskip.5\tabcolsep}
\newcommand{\cdashlinelr}[1]{%
  \noalign{\vskip\aboverulesep
           \global\let\@dashdrawstore\adl@draw
           \global\let\adl@draw\adl@drawiv}
  \cdashline{#1}
  \noalign{\global\let\adl@draw\@dashdrawstore
           \vskip\belowrulesep}}
\makeatother

\newcommand{\ourmodel}{RoBERTa}
\newcommand{\ourmodelbase}{RoBERTa$_{\textsc{base}}$}
\newcommand{\ourmodellarge}{RoBERTa$_{\textsc{large}}$}
\newcommand{\bertbase}{BERT$_{\textsc{base}}$}
\newcommand{\bertlarge}{BERT$_{\textsc{large}}$}
\newcommand{\xlnetbase}{XLNet$_{\textsc{base}}$}
\newcommand{\xlnetlarge}{XLNet$_{\textsc{large}}$}

\title{\ourmodel{}: A Robustly Optimized BERT Pretraining Approach}

\author{Yinhan Liu\thanks{~~Equal contribution.} $^{\mathsection}$ \quad Myle Ott$^{*\mathsection}$ \quad Naman Goyal$^{* \mathsection}$ \quad Jingfei Du$^{* \mathsection}$ \quad Mandar Joshi$^{\dagger}$\\
{ \bf Danqi Chen$^{\mathsection}$ \quad Omer Levy$^{\mathsection}$ \quad Mike Lewis$^{\mathsection}$ \quad Luke Zettlemoyer$^{\dagger\mathsection}$\quad Veselin Stoyanov$^{\mathsection}$} \\[8pt]
$^{\dagger}$ Paul G. Allen School of Computer Science \& Engineering, \\ University of Washington, Seattle, WA \\
{\tt \{mandar90,lsz\}@cs.washington.edu}\\[4pt]
$^{\mathsection}$ Facebook AI \\
{\tt \{yinhanliu,myleott,naman,jingfeidu,}\\
{\tt \quad\quad\quad\quad danqi,omerlevy,mikelewis,lsz,ves\}@fb.com}
}
\date{}

\begin{document}

\maketitle

\begin{abstract}

Language model pretraining has led to significant performance gains but careful comparison between different approaches is challenging.
Training is computationally expensive, often done on private datasets of different sizes, and, as we will show, hyperparameter choices have significant impact on the final results. 
We present a replication study of BERT pretraining~\cite{devlin2018bert} that carefully measures the impact of many key hyperparameters and training data size. We find that BERT was significantly undertrained, and can match or exceed the performance of every model published after it. Our best model achieves state-of-the-art results on GLUE, RACE and SQuAD.
These results highlight the importance of previously overlooked design choices, and raise questions about the source of recently reported improvements. We release our models and code.\footnote{Our models and code are available at: \\
\url{https://github.com/pytorch/fairseq}}

\end{abstract}
\section{Introduction}
\label{intro}

Self-training methods such as ELMo~\cite{peters2018deep}, GPT~\cite{radford2018gpt}, BERT \cite{devlin2018bert}, XLM~\cite{lample2019cross}, and XLNet \cite{yang2019xlnet} have brought significant performance gains, but it can be challenging to determine which aspects of the methods contribute the most. %
Training is computationally expensive, limiting the amount of tuning that can be done, and is often done with private training data of varying sizes, limiting our ability to measure the effects of the modeling advances.

We present a replication study of BERT pretraining~\cite{devlin2018bert}, which includes a careful evaluation of the effects of hyperparmeter tuning and training set size. %
We find that BERT was significantly undertrained and propose an improved recipe for training BERT models, which we call \ourmodel{}, that can match or exceed the performance of all of the post-BERT methods.
Our modifications are simple, they include: (1) training the model longer, with bigger batches, over more data; (2) removing the next sentence prediction objective; (3) training on longer sequences; and (4) dynamically changing the masking pattern applied to the training data. We also collect a large new dataset (\textsc{CC-News}) of comparable size to other privately used datasets, to better control for training set size effects. 

When controlling for training data, our improved training procedure improves upon the published BERT results on both GLUE and SQuAD.
When trained for longer over additional data, our model achieves a score of 88.5 on the public GLUE leaderboard, matching the 88.4 reported by \newcite{yang2019xlnet}.
Our model establishes a new state-of-the-art on 4/9 of the GLUE tasks: MNLI, QNLI, RTE and STS-B.
We also match state-of-the-art results on SQuAD and RACE.
Overall, we re-establish that BERT's masked language model training objective is competitive with other recently proposed training objectives such as perturbed autoregressive language modeling~\cite{yang2019xlnet}.\footnote{It is possible that these other methods could also improve with more tuning. We leave this exploration to future work.}

In summary, the contributions of this paper are: (1) We present a set of important BERT design choices and training strategies and introduce alternatives that lead to better downstream task performance; (2) We use a novel dataset, \textsc{CC-News}, and confirm that using more data for pretraining further improves performance on downstream tasks; (3) Our training improvements show that masked language model pretraining, under the right design choices, is competitive with all other recently published methods. We release our model, pretraining and fine-tuning code implemented in PyTorch~\cite{paszke2017automatic}.
\section{Background} \label{sec:background}

In this section, we give a brief overview of the BERT~\cite{devlin2018bert} pretraining approach and some of the training choices that we will examine experimentally in the following section.

\subsection{Setup}
BERT takes as input a concatenation of two segments (sequences of tokens), $x_1 , \ldots , x_N$ and $y_1 , \ldots , y_M$.
Segments usually consist of more than one natural sentence.
The two segments are presented as a single input sequence to BERT with special tokens delimiting them: $[\mathit{CLS}], x_1 , \ldots , x_N, [\mathit{SEP}], y_1 , \ldots , y_M, [\mathit{EOS}]$.
$M$ and $N$ are constrained such that $M + N < T$, where $T$ is a parameter that controls the maximum sequence length during training.

The model is first pretrained on a large unlabeled text corpus and subsequently finetuned using end-task labeled data.

\subsection{Architecture}
BERT uses the now ubiquitous transformer architecture \cite{vaswani2017attention}, which we will not review in detail. We use a transformer architecture with $L$ layers. Each block uses $A$ self-attention heads and hidden dimension $H$.

\subsection{Training Objectives}

During pretraining, BERT uses two objectives: masked language modeling and next sentence prediction. 

\paragraph{Masked Language Model (MLM)} A random sample of the tokens in the input sequence is selected and replaced with the special token $[\mathit{MASK}]$. The MLM objective is a cross-entropy loss on predicting the masked tokens. BERT uniformly selects 15\% of the input tokens for possible replacement. Of the selected tokens, 80\% are replaced with $[\mathit{MASK}]$, 10\% are left unchanged, and 10\% are replaced by a randomly selected vocabulary token. 
    
In the original implementation, random masking and replacement is performed once in the beginning and saved for the duration of training, although in practice, data is duplicated so the mask is not always the same for every training sentence (see Section \ref{sec:dynamic_masking}).

\paragraph{Next Sentence Prediction (NSP)} NSP is a binary classification loss for predicting whether two segments follow each other in the original text. Positive examples are created by taking consecutive sentences from the text corpus. Negative examples are created by pairing segments from different documents. Positive and negative examples are sampled with equal probability. 
    
The NSP objective was designed to improve performance on downstream tasks, such as Natural Language Inference \cite{bowman2015large}, which require reasoning about the relationships between pairs of sentences.

\subsection{Optimization}

BERT is optimized with Adam \cite{kingma2014adam} using the following parameters: $\beta_1 = 0.9$, $\beta_2= 0.999$, $\epsilon = \text{1e-6}$ and $L_2$ weight decay of $0.01$. The learning rate is warmed up over the first 10,000 steps to a peak value of 1e-4, and then linearly decayed. BERT trains with a dropout of 0.1 on all layers and attention weights, and a GELU activation function~\cite{hendrycks2016gelu}. Models are pretrained for $S=\text{1,000,000}$ updates, with minibatches containing $B=\text{256}$ sequences of maximum length $T=\text{512}$ tokens.

\subsection{Data}

BERT is trained on a combination of \textsc{BookCorpus}~\cite{moviebook} plus English \textsc{Wikipedia}, which totals 16GB of uncompressed text.\footnote{\newcite{yang2019xlnet} use the same dataset but report having only 13GB of text after data cleaning. This is most likely due to subtle differences in cleaning of the Wikipedia data.}

\section{Experimental Setup} \label{sec:exp}

In this section, we describe the experimental setup for our replication study of BERT.

\subsection{Implementation} \label{sec:implementation}

We reimplement BERT in \textsc{fairseq}~\cite{ott2019fairseq}.
We primarily follow the original BERT optimization hyperparameters, given in Section~\ref{sec:background}, except for the peak learning rate and number of warmup steps, which are tuned separately for each setting.
We additionally found training to be very sensitive to the Adam epsilon term, and in some cases we obtained better performance or improved stability after tuning it.
Similarly, we found setting $\beta_2 = 0.98$ to improve stability when training with large batch sizes.

We pretrain with sequences of at most $T=512$ tokens.
Unlike \newcite{devlin2018bert}, we do not randomly inject short sequences, and we do not train with a reduced sequence length for the first 90\% of updates.
We train only with full-length sequences.

We train with mixed precision floating point arithmetic on DGX-1 machines, each with 8 $\times$ 32GB Nvidia V100 GPUs interconnected by Infiniband~\cite{micikevicius2018mixed}.

\subsection{Data} \label{sec:data}

BERT-style pretraining crucially relies on large quantities of text. \newcite{baevski2019cloze} demonstrate that increasing data size can result in improved end-task performance. Several efforts have trained on datasets larger and more diverse than the original BERT~\cite{radford2019language,yang2019xlnet,zellers2019neuralfakenews}.
Unfortunately, not all of the additional datasets can be publicly released. For our study, we focus on gathering as much data as possible for experimentation, allowing us to match the overall quality and quantity of data as appropriate for each comparison. 

We consider five English-language corpora of varying sizes and domains, totaling over 160GB of uncompressed text. We use the following text corpora:
\begin{itemize}[leftmargin=*]
\setlength\itemsep{0em}
\item \textsc{BookCorpus}~\cite{moviebook} plus English \textsc{Wikipedia}. This is the original data used to train BERT. (16GB).
\item \textsc{CC-News}, which we collected from the English portion of the CommonCrawl News dataset~\cite{nagel2016ccnews}. The data contains 63 million English news articles crawled between September 2016 and February 2019. (76GB after filtering).\footnote{We use \texttt{news-please}~\cite{hamborg2017newsplease} to collect and extract \textsc{CC-News}. \textsc{CC-News} is similar to the \textsc{RealNews} dataset described in~\newcite{zellers2019neuralfakenews}.}
\item \textsc{OpenWebText}~\cite{gokaslan2019openwebtext}, an open-source recreation of the WebText corpus described in~\newcite{radford2019language}. The text is web content extracted from URLs shared on Reddit with at least three upvotes. (38GB).\footnote{The authors and their affiliated institutions are not in any way affiliated with the creation of the OpenWebText dataset.}
\item \textsc{Stories}, a dataset introduced in~\newcite{trinh2018simple} containing a subset of CommonCrawl data filtered to match the story-like style of Winograd schemas. (31GB).
\end{itemize}

\subsection{Evaluation} \label{sec:evaluation}

Following previous work, we evaluate our pretrained models on downstream tasks using the following three benchmarks.

\paragraph{GLUE} \label{sec:glue}

The General Language Understanding Evaluation (GLUE) benchmark \cite{wang2019glue} is a collection of 9 datasets for evaluating natural language understanding systems.\footnote{The datasets are: CoLA~\cite{warstadt2018neural}, Stanford Sentiment Treebank (SST)~\cite{socher2013recursive}, Microsoft Research Paragraph Corpus (MRPC)~\cite{dolan2005automatically}, Semantic Textual Similarity Benchmark (STS)~\cite{agirre2007semantic}, Quora Question Pairs (QQP)~\cite{iyer2016quora}, Multi-Genre NLI (MNLI)~\cite{williams2018broad}, Question NLI (QNLI)~\cite{rajpurkar2016squad}, Recognizing Textual Entailment (RTE)~\cite{dagan2006pascal,bar2006second,giampiccolo2007third,bentivogli2009fifth} and Winograd NLI (WNLI)~\cite{levesque2011winograd}.}
Tasks are framed as either single-sentence classification or sentence-pair classification tasks.
The GLUE organizers provide training and development data splits as well as a submission server and leaderboard that allows participants to evaluate and compare their systems on private held-out test data.

For the replication study in Section~\ref{sec:design}, we report results on the development sets after finetuning the pretrained models on the corresponding single-task training data (i.e., without multi-task training or ensembling).
Our finetuning procedure follows the original BERT paper~\cite{devlin2018bert}.

In Section~\ref{sec:roberta} we additionally report test set results obtained from the public leaderboard. These results depend on a several task-specific modifications, which we describe in Section~\ref{sec:results_glue}.

\paragraph{SQuAD} \label{sec:squad}
The Stanford Question Answering Dataset (SQuAD) provides a paragraph of context and a  question.
The task is to answer the question by extracting the relevant span from the context.
We evaluate on two versions of SQuAD: V1.1 and V2.0~\cite{rajpurkar2016squad,rajpurkar2018know}.
In V1.1 the context always contains an answer, whereas in V2.0 some questions are not answered in the provided context, making the task more challenging.

For SQuAD V1.1 we adopt the same span prediction method as BERT~\cite{devlin2018bert}.
For SQuAD V2.0, we add an additional binary classifier to predict whether the question is answerable, which we train jointly by summing the classification and span loss terms.
During evaluation, we only predict span indices on pairs that are classified as answerable.

\paragraph{RACE} \label{sec:race}
The ReAding Comprehension from Examinations (RACE)~\cite{lai2017large} task is a large-scale reading comprehension dataset with more than 28,000 passages and nearly 100,000 questions. The dataset is collected from English examinations in China, which are designed for middle and high school students. In RACE, each passage is associated with multiple questions. For every question, the task is to select one correct answer from four options. RACE has significantly longer context than other popular reading comprehension datasets and the proportion of questions
that requires reasoning is very large.
\section{Training Procedure Analysis} \label{sec:design}

This section explores and quantifies which choices are important for successfully pretraining BERT models.
We keep the model architecture fixed.\footnote{Studying architectural changes, including larger architectures, is an important area for future work.} 
Specifically, we begin by training BERT models with the same configuration as BERT$_{\textsc{base}}$ ($L=12$,
$H=768$, $A=12$, 110M params).

\subsection{Static vs. Dynamic Masking} \label{sec:dynamic_masking}

As discussed in Section \ref{sec:background}, BERT relies on randomly masking and predicting tokens. 
The original BERT implementation performed masking once during data preprocessing, resulting in a single \emph{static} mask.
To avoid using the same mask for each training instance in every epoch, training data was duplicated 10 times so that each sequence is masked in 10 different ways over the 40 epochs of training.
Thus, each training sequence was seen with the same mask four times during training.

We compare this strategy with \emph{dynamic masking} where we generate the masking pattern every time we feed a sequence to the model.
This becomes crucial when pretraining for more steps or with larger datasets.

\paragraph{Results}

\begin{table}[t]
\begin{center}
\begin{tabular}{lccc}
\toprule
\bf Masking & \bf SQuAD 2.0 & \bf MNLI-m & \bf SST-2 \\
\midrule
reference & 76.3 & 84.3 & 92.8 \\
\midrule
\multicolumn{4}{l}{\emph{Our reimplementation:}} \\
static & 78.3 & 84.3 & 92.5 \\
dynamic & 78.7 & 84.0 & 92.9 \\
\bottomrule
\end{tabular}
\end{center}
\caption{Comparison between static and dynamic masking for \bertbase{}.
We report F1 for SQuAD and accuracy for MNLI-m and SST-2.
Reported results are medians over 5 random initializations (seeds).
Reference results are from \newcite{yang2019xlnet}.}
\label{tab:static_vs_dynamic_masking}
\end{table}

Table~\ref{tab:static_vs_dynamic_masking} compares the published \bertbase{} results from \newcite{devlin2018bert} to our reimplementation with either static or dynamic masking.
We find that our reimplementation with static masking performs similar to the original BERT model, and dynamic masking is comparable or slightly better than static masking.

Given these results and the additional efficiency benefits of dynamic masking, we use dynamic masking in the remainder of the experiments.

\subsection{Model Input Format and Next Sentence Prediction} \label{sec:model_input_nsp}

\begin{table*}[t]
\begin{center}
\begin{tabular}{lcccc}
\toprule
\bf Model & \bf SQuAD 1.1/2.0 & \bf MNLI-m & \bf SST-2 & \bf RACE \\ 
\midrule
\multicolumn{5}{l}{\emph{Our reimplementation (with NSP loss):}} \\
\textsc{segment-pair} & 90.4/78.7 & 84.0 & 92.9 & 64.2 \\
\textsc{sentence-pair} & 88.7/76.2 & 82.9 & 92.1 & 63.0 \\
\midrule
\multicolumn{5}{l}{\emph{Our reimplementation (without NSP loss):}} \\
\textsc{full-sentences} & 90.4/79.1 & 84.7 & 92.5 & 64.8 \\
\textsc{doc-sentences} & 90.6/79.7 & 84.7 & 92.7 & 65.6 \\
\midrule
\bertbase{} & 88.5/76.3 & 84.3 & 92.8 & 64.3 \\
\xlnetbase{} (K = 7) & --/81.3 & 85.8 & 92.7 & 66.1 \\
\xlnetbase{} (K = 6) & --/81.0 & 85.6 & 93.4 & 66.7 \\
\bottomrule

\end{tabular}
\end{center}
\caption{
Development set results for base models pretrained over \textsc{BookCorpus} and \textsc{Wikipedia}.
All models are trained for 1M steps with a batch size of 256 sequences.
We report F1 for SQuAD and accuracy for MNLI-m, SST-2 and RACE.
Reported results are medians over five random initializations (seeds).
Results for \bertbase{} and \xlnetbase{} are from \newcite{yang2019xlnet}.
}
\label{tab:base_apples_to_apples}
\end{table*}

In the original BERT pretraining procedure, the model observes two concatenated document segments, which are either sampled contiguously from the same document (with $p=0.5$) or from distinct documents.
In addition to the masked language modeling objective, the model is trained to predict whether the observed document segments come from the same or distinct documents via an auxiliary Next Sentence Prediction (NSP) loss.

The NSP loss was hypothesized to be an important factor in training the original BERT model. \newcite{devlin2018bert} observe that removing NSP hurts performance, with significant performance degradation on QNLI, MNLI, and SQuAD 1.1.
However, some recent work has questioned the necessity of the NSP loss~\cite{lample2019cross,yang2019xlnet,joshi2019spanbert}.

To better understand this discrepancy, we compare several alternative training formats:
\begin{itemize}[leftmargin=*]
\setlength\itemsep{0em}
\item \textsc{segment-pair+nsp}: This follows the original input format used in BERT~\cite{devlin2018bert}, with the NSP loss. Each input has a pair of segments, which can each contain multiple natural sentences, but the total combined length must be less than 512 tokens.
\item \textsc{sentence-pair+nsp}: Each input contains a pair of natural \emph{sentences}, either sampled from a contiguous portion of one document or from separate documents. Since these inputs are significantly shorter than 512 tokens, we increase the batch size so that the total number of tokens remains similar to \textsc{segment-pair+nsp}. We retain the NSP loss.
\item \textsc{full-sentences}: Each input is packed with full sentences sampled contiguously from one or more documents, such that the total length is at most 512 tokens. Inputs may cross document boundaries. When we reach the end of one document, we begin sampling sentences from the next document and add an extra separator token between documents. We remove the NSP loss.
\item \textsc{doc-sentences}: Inputs are constructed similarly to \textsc{full-sentences}, except that they may not cross document boundaries. Inputs sampled near the end of a document may be shorter than 512 tokens, so we dynamically increase the batch size in these cases to achieve a similar number of total tokens as \textsc{full-sentences}. We remove the NSP loss.
\end{itemize}

\paragraph{Results}

Table~\ref{tab:base_apples_to_apples} shows results for the four different settings.
We first compare the original \textsc{segment-pair} input format from \newcite{devlin2018bert} to the \textsc{sentence-pair} format; both formats retain the NSP loss, but the latter uses single sentences.
We find that \textbf{using individual sentences hurts performance on downstream tasks}, which we hypothesize is because the model is not able to learn long-range dependencies.

We next compare training without the NSP loss and training with blocks of text from a single document (\textsc{doc-sentences}).
We find that this setting outperforms the originally published \bertbase{} results and that \textbf{removing the NSP loss matches or slightly improves downstream task performance}, in contrast to \newcite{devlin2018bert}.
It is possible that the original BERT implementation may only have removed the loss term while still retaining the \textsc{segment-pair} input format.

Finally we find that restricting sequences to come from a single document (\textsc{doc-sentences}) performs slightly better than packing sequences from multiple documents (\textsc{full-sentences}).
However, because the \textsc{doc-sentences} format results in variable batch sizes, we use \textsc{full-sentences} in the remainder of our experiments for easier comparison with related work.

\subsection{Training with large batches}
\label{sec:large_batches}

Past work in Neural Machine Translation has shown that training with very large mini-batches can both improve optimization speed and end-task performance when the learning rate is increased appropriately~\cite{ott2018scaling}.
Recent work has shown that BERT is also amenable to large batch training~\cite{you2019reducing}.

\newcite{devlin2018bert} originally trained \bertbase{} for 1M steps with a batch size of 256 sequences.
This is equivalent in computational cost, via gradient accumulation, to training for 125K steps with a batch size of 2K sequences, or for 31K steps with a batch size of 8K.

\begin{table}[t]

\begin{center}
\begin{tabular}{cccccc}
\toprule

\bf bsz & \bf steps & \bf lr & \bf ppl & \bf MNLI-m & \bf SST-2 \\
\midrule
256 & 1M & 1e-4 & 3.99 & 84.7 & 92.7 \\
2K & 125K & 7e-4 & \textbf{3.68} & \textbf{85.2} & \textbf{92.9} \\
8K & 31K & 1e-3 & 3.77 & 84.6 & 92.8 \\
\bottomrule
\end{tabular}
\end{center}
\caption{
Perplexity on held-out training data (\emph{ppl}) and development set accuracy for base models trained over \textsc{BookCorpus} and \textsc{Wikipedia} with varying batch sizes (\emph{bsz}).
We tune the learning rate (\emph{lr}) for each setting.
Models make the same number of passes over the data (epochs) and have the same computational cost.
}

\label{tab:large_batches}

\end{table}

In Table~\ref{tab:large_batches} we compare perplexity and end-task performance of \bertbase{} as we increase the batch size, controlling for the number of passes through the training data.
We observe that training with large batches improves perplexity for the masked language modeling objective, as well as end-task accuracy.
Large batches are also easier to parallelize via distributed data parallel training,\footnote{Large batch training can improve training efficiency even without large scale parallel hardware through \emph{gradient accumulation}, whereby gradients from multiple mini-batches are accumulated locally before each optimization step. This functionality is supported natively in \textsc{fairseq}~\cite{ott2019fairseq}.} and in later experiments we train with batches of 8K sequences.

Notably \newcite{you2019reducing} train BERT with even larger batche sizes, up to 32K sequences.
We leave further exploration of the limits of large batch training to future work.

\subsection{Text Encoding}
\label{sec:bpe}

Byte-Pair Encoding (BPE)~\cite{sennrich2016neural} is a hybrid between character- and word-level representations that allows handling the large vocabularies common in natural language corpora.
Instead of full words, BPE relies on subwords units, which are extracted by performing statistical analysis of the training corpus.

BPE vocabulary sizes typically range from 10K-100K subword units. However, unicode characters can account for a sizeable portion of this vocabulary when modeling large and diverse corpora, such as the ones considered in this work.
\newcite{radford2019language} introduce a clever implementation of BPE that uses \emph{bytes} instead of unicode characters as the base subword units.
Using bytes makes it possible to learn a subword vocabulary of a modest size (50K units) that can still encode any input text without introducing any ``unknown" tokens.

The original BERT implementation~\cite{devlin2018bert} uses a character-level BPE vocabulary of size 30K, which is learned after preprocessing the input with heuristic tokenization rules.
Following \newcite{radford2019language}, we instead consider training BERT with a larger byte-level BPE vocabulary containing 50K subword units, without any additional preprocessing or tokenization of the input.
This adds approximately 15M and 20M additional parameters for \bertbase{} and \bertlarge{}, respectively.

Early experiments revealed only slight differences between these encodings, with the \newcite{radford2019language} BPE achieving slightly worse end-task performance on some tasks.
Nevertheless, we believe the advantages of a universal encoding scheme outweighs the minor degredation in performance and use this encoding in the remainder of our experiments.
A more detailed comparison of these encodings is left to future work.
\section{\ourmodel{}} \label{sec:roberta}

\begin{table*}[t]
\begin{center}
\begin{tabular}{lcccccc}
\toprule
\multirow{2}{*}{\bf Model} & \bf \multirow{2}{*}{data} & \bf \multirow{2}{*}{bsz} & \bf \multirow{2}{*}{steps} & \bf SQuAD & \bf \multirow{2}{*}{MNLI-m} & \bf \multirow{2}{*}{SST-2} \\
& & & & (v1.1/2.0) & & \\
\midrule
\multicolumn{4}{l}{\ourmodel{}} \\
\quad with \textsc{Books} + \textsc{Wiki} & 16GB & 8K & 100K & 93.6/87.3 & 89.0 & 95.3 \\
\quad + additional data (\textsection\ref{sec:data}) & 160GB & 8K & 100K & 94.0/87.7 & 89.3 & 95.6 \\
\quad + pretrain longer & 160GB & 8K & 300K & 94.4/88.7 & 90.0 & 96.1 \\
\quad + pretrain even longer & 160GB & 8K & 500K & \textbf{94.6}/\textbf{89.4} & \textbf{90.2} & \textbf{96.4} \\
\midrule
\multicolumn{4}{l}{\bertlarge{}} \\
\quad with \textsc{Books} + \textsc{Wiki} & 13GB & 256 & 1M & 90.9/81.8 & 86.6 & 93.7 \\
\multicolumn{4}{l}{\xlnetlarge{}} \\
\quad with \textsc{Books} + \textsc{Wiki} & 13GB & 256 & 1M & 94.0/87.8 & 88.4 & 94.4 \\
\quad + additional data & 126GB & 2K & 500K & 94.5/88.8 & 89.8 & 95.6 \\
\bottomrule
\end{tabular}
\end{center}
\caption{Development set results for \ourmodel{} as we pretrain over more data (16GB $\rightarrow$ 160GB of text) and pretrain for longer (100K $\rightarrow$ 300K $\rightarrow$ 500K steps).
Each row accumulates improvements from the rows above.
\ourmodel{} matches the architecture and training objective of \bertlarge{}.
Results for \bertlarge{} and \xlnetlarge{} are from \newcite{devlin2018bert} and \newcite{yang2019xlnet}, respectively.
Complete results on all GLUE tasks can be found in the Appendix.
}
\label{tab:ablation}
\end{table*}

In the previous section we propose modifications to the BERT pretraining procedure that improve end-task performance.
We now aggregate these improvements and evaluate their combined impact.
We call this configuration \textbf{\ourmodel{}} for \underline{\textbf{R}}obustly \underline{\textbf{o}}ptimized \underline{\textbf{BERT}} \underline{\textbf{a}}pproach.
Specifically, \ourmodel{} is trained with dynamic masking (Section~\ref{sec:dynamic_masking}), \textsc{full-sentences} without NSP loss (Section~\ref{sec:model_input_nsp}), large mini-batches (Section~\ref{sec:large_batches}) and a larger byte-level BPE (Section~\ref{sec:bpe}).

Additionally, we investigate two other important factors that have been under-emphasized in previous work: (1) the data used for pretraining, and (2) the number of training passes through the data.
For example, the recently proposed XLNet architecture~\cite{yang2019xlnet} is pretrained using nearly 10 times more data than the original BERT~\cite{devlin2018bert}.
It is also trained with a batch size eight times larger for half as many optimization steps, thus seeing four times as many sequences in pretraining compared to BERT.

To help disentangle the importance of these factors from other modeling choices (e.g., the pretraining objective), we begin by training \ourmodel{} following the \bertlarge{} architecture ($L=24$, $H=1024$, $A=16$, 355M parameters).
We pretrain for 100K steps over a comparable \textsc{BookCorpus} plus \textsc{Wikipedia} dataset as was used in \newcite{devlin2018bert}.
We pretrain our model using 1024 V100 GPUs for approximately one day.

\paragraph{Results}

We present our results in Table~\ref{tab:ablation}.
When controlling for training data, we observe that \ourmodel{} provides a large improvement over the originally reported \bertlarge{} results, reaffirming the importance of the design choices we explored in Section~\ref{sec:design}.

Next, we combine this data with the three additional datasets described in Section~\ref{sec:data}.
We train \ourmodel{} over the combined data with the same number of training steps as before (100K).
In total, we pretrain over 160GB of text.
We observe further improvements in performance across all downstream tasks, validating the importance of data size and diversity in pretraining.\footnote{Our experiments conflate increases in data size and diversity. We leave a more careful analysis of these two dimensions to future work.}

\begin{table*}[t]
\begin{center}
\begin{tabular}{lcccccccccc}
\toprule
& \bf MNLI & \bf QNLI & \bf QQP & \bf RTE & \bf SST & \bf MRPC & \bf CoLA & \bf STS & \bf WNLI & \bf Avg \\
\midrule 
\multicolumn{10}{l}{\textit{Single-task single models on dev}}\\
\bertlarge{} & 86.6/- & 92.3 & 91.3 & 70.4 & 93.2 & 88.0 & 60.6 & 90.0 & - & -\\
\xlnetlarge{} & 89.8/- & 93.9 & 91.8 & 83.8 & 95.6 & 89.2 & 63.6 & 91.8 & - & -\\
\ourmodel{} & \textbf{90.2}/\textbf{90.2} & \textbf{94.7} & \textbf{92.2} & \textbf{86.6} & \textbf{96.4} & \textbf{90.9} & \textbf{68.0} & \textbf{92.4} & \textbf{91.3} & - \\
\midrule
\multicolumn{10}{l}{\textit{Ensembles on test (from leaderboard as of July 25, 2019)}} \\
ALICE & 88.2/87.9 & 95.7 & \textbf{90.7} & 83.5 & 95.2 & 92.6 & \textbf{68.6} & 91.1 & 80.8 & 86.3 \\
MT-DNN & 87.9/87.4 & 96.0 & 89.9 & 86.3 & 96.5 & 92.7 & 68.4 & 91.1 & 89.0 & 87.6 \\
XLNet  & 90.2/89.8 & 98.6 & 90.3 & 86.3 & \textbf{96.8} & \textbf{93.0} & 67.8 & 91.6 & \textbf{90.4} & 88.4 \\
\ourmodel{} & \textbf{90.8/90.2} & \textbf{98.9} & 90.2 & \textbf{88.2} & 96.7 & 92.3 & 67.8 & \textbf{92.2} & 89.0 & \bf 88.5 \\
\bottomrule
\end{tabular}
\end{center}
\caption{
Results on GLUE. All results are based on a 24-layer architecture.
\bertlarge{} and \xlnetlarge{} results are from \newcite{devlin2018bert} and \newcite{yang2019xlnet}, respectively.
\ourmodel{} results on the development set are a median over five runs.
\ourmodel{} results on the test set are ensembles of \emph{single-task} models.
For RTE, STS and MRPC we finetune starting from the MNLI model instead of the baseline pretrained model.
Averages are obtained from the GLUE leaderboard.
}
\label{tab:roberta_glue}
\end{table*}

Finally, we pretrain \ourmodel{} for significantly longer, increasing the number of pretraining steps from 100K to 300K, and then further to 500K.
We again observe significant gains in downstream task performance, and the 300K and 500K step models outperform \xlnetlarge{} across most tasks.
We note that even our longest-trained model does not appear to overfit our data and would likely benefit from additional training.

In the rest of the paper, we evaluate our best \ourmodel{} model on the three different benchmarks: GLUE, SQuaD and RACE.
Specifically we consider \ourmodel{} trained for 500K steps over all five of the datasets introduced in Section~\ref{sec:data}.

\subsection{GLUE Results} \label{sec:results_glue}

For GLUE we consider two finetuning settings.
In the first setting (\emph{single-task, dev}) we finetune \ourmodel{} separately for each of the GLUE tasks, using only the training data for the corresponding task.
We consider a limited hyperparameter sweep for each task, with batch sizes $\in \{16, 32\}$ and learning rates $\in \{1e-5, 2e-5, 3e-5\}$, with a linear warmup for the first 6\% of steps followed by a linear decay to 0.
We finetune for 10 epochs and perform early stopping based on each task's evaluation metric on the dev set.
The rest of the hyperparameters remain the same as during pretraining.
In this setting, we report the median development set results for each task over five random initializations, without model ensembling.

In the second setting (\emph{ensembles, test}), we compare \ourmodel{} to other approaches on the test set via the GLUE leaderboard.
While many submissions to the GLUE leaderboard depend on multi-task finetuning, \textbf{our submission depends only on single-task finetuning}.
For RTE, STS and MRPC we found it helpful to finetune starting from the MNLI single-task model, rather than the baseline pretrained \ourmodel{}.
We explore a slightly wider hyperparameter space, described in the Appendix, and ensemble between 5 and 7 models per task.

\paragraph{Task-specific modifications}

Two of the GLUE tasks require task-specific finetuning approaches to achieve competitive leaderboard results.

\underline{QNLI}:
Recent submissions on the GLUE leaderboard adopt a pairwise ranking formulation for the QNLI task, in which candidate answers are mined from the training set and compared to one another, and a single (question, candidate) pair is classified as positive~\cite{liu2019mtdnn,liu2019improving,yang2019xlnet}.
This formulation significantly simplifies the task, but is not directly comparable to BERT~\cite{devlin2018bert}.
Following recent work, we adopt the ranking approach for our test submission, but for direct comparison with BERT we report development set results based on a pure classification approach.

\underline{WNLI}: We found the provided NLI-format data to be challenging to work with.
Instead we use the reformatted WNLI data from SuperGLUE~\cite{wang2019superglue}, which indicates the span of the query pronoun and referent.
We finetune \ourmodel{} using the margin ranking loss from \newcite{kocijan2019surprisingly}.
For a given input sentence, we use spaCy~\cite{spacy2} to extract additional candidate noun phrases from the sentence and finetune our model so that it assigns higher scores to positive referent phrases than for any of the generated negative candidate phrases.
One unfortunate consequence of this formulation is that we can only make use of the positive training examples, which excludes over half of the provided training examples.\footnote{While we only use the provided WNLI training data, our results could potentially be improved by augmenting this with additional pronoun disambiguation datasets.}

\paragraph{Results}

We present our results in Table~\ref{tab:roberta_glue}.
In the first setting (\emph{single-task, dev}), \ourmodel{} achieves state-of-the-art results on all 9 of the GLUE task development sets.
Crucially, \ourmodel{} uses the same masked language modeling pretraining objective and architecture as \bertlarge{}, yet consistently outperforms both \bertlarge{} and \xlnetlarge{}.
This raises questions about the relative importance of model architecture and pretraining objective, compared to more mundane details like dataset size and training time that we explore in this work.

In the second setting (\emph{ensembles, test}), we submit \ourmodel{} to the GLUE leaderboard and achieve state-of-the-art results on 4 out of 9 tasks and the highest average score to date.
This is especially exciting because \ourmodel{} does not depend on multi-task finetuning, unlike most of the other top submissions.
We expect future work may further improve these results by incorporating more sophisticated multi-task finetuning procedures.

\subsection{SQuAD Results} \label{sec:results_squad}

We adopt a much simpler approach for SQuAD compared to past work.
In particular, while both BERT~\cite{devlin2018bert} and XLNet~\cite{yang2019xlnet} augment their training data with additional QA datasets, \textbf{we only finetune \ourmodel{} using the provided SQuAD training data}.
\newcite{yang2019xlnet} also employed a custom layer-wise learning rate schedule to finetune XLNet, while we use the same learning rate for all layers.

For SQuAD v1.1 we follow the same finetuning procedure as \newcite{devlin2018bert}.
For SQuAD v2.0, we additionally classify whether a given question is answerable; we train this classifier jointly with the span predictor by summing the classification and span loss terms.

\paragraph{Results}

\begin{table}[t]
\begin{center}
\begin{tabular}{lcccc}
\toprule
\multirow{2}{*}{\bf Model} & \multicolumn{2}{c}{\bf SQuAD 1.1} &\multicolumn{2}{c}{\bf SQuAD 2.0} \\
&  EM &  F1 &  EM &  F1  \\
\midrule
\multicolumn{5}{l}{\textit{Single models on dev, w/o data augmentation}}\\
\bertlarge{} &  84.1&90.9&79.0&81.8\\
\xlnetlarge{} &\bf{89.0}& 94.5&86.1&88.8\\
\ourmodel{} & 88.9 & \bf{94.6} & \bf{86.5} &\bf{89.4}\\
\midrule
\multicolumn{5}{l}{\textit{Single models on test (as of July 25, 2019)}}\\
\multicolumn{3}{l}{\xlnetlarge{}} & 86.3$^{\dag}$ & 89.1$^{\dag}$ \\
\multicolumn{3}{l}{\ourmodel{}} & 86.8 & 89.8 \\
\multicolumn{3}{l}{XLNet + SG-Net Verifier} & \textbf{87.0}$^{\dag}$ & \textbf{89.9}$^{\dag}$ \\
\bottomrule
\end{tabular}
\end{center}
\caption{
Results on SQuAD.
$\dag$ indicates results that depend on additional external training data.
\ourmodel{} uses only the provided SQuAD data in both dev and test settings.
BERT$_{\textsc{large}}$ and XLNet$_{\textsc{large}}$ results are from \newcite{devlin2018bert} and \newcite{yang2019xlnet}, respectively.
}
\label{tab:roberta_squad}
\end{table}

We present our results in Table~\ref{tab:roberta_squad}.
On the SQuAD v1.1 development set, \ourmodel{} matches the state-of-the-art set by XLNet.
On the SQuAD v2.0 development set, \ourmodel{} sets a new state-of-the-art, improving over XLNet by 0.4 points (EM) and 0.6 points (F1).

We also submit \ourmodel{} to the public SQuAD 2.0 leaderboard and evaluate its performance relative to other systems.
Most of the top systems build upon either BERT~\cite{devlin2018bert} or XLNet~\cite{yang2019xlnet}, both of which rely on additional external training data.
In contrast, our submission does not use any additional data.

Our single \ourmodel{} model outperforms all but one of the single model submissions, and is the top scoring system among those that do not rely on data augmentation.

\subsection{RACE Results} \label{sec:results_race}

In RACE, systems are provided with a passage of text, an associated question, and four candidate answers. Systems are required to classify which of the four candidate answers is correct.

We modify \ourmodel{} for this task by concatenating each candidate answer with the corresponding question and passage.
We then encode each of these four sequences and pass the resulting \emph{[CLS]} representations through a fully-connected layer, which is used to predict the correct answer.
We truncate question-answer pairs that are longer than 128 tokens and, if needed, the passage so that the total length is at most 512 tokens.

\begin{table}[t]
\begin{center}
\begin{tabular}{lccc}
\toprule
\bf Model & \bf Accuracy & \bf Middle & \bf High \\
\midrule
\multicolumn{4}{l}{\textit{Single models on test (as of July 25, 2019)}}\\
\bertlarge{} &  72.0 & 76.6 & 70.1 \\
\xlnetlarge{} & 81.7 & 85.4 & 80.2 \\
\midrule
\ourmodel{} & \bf{83.2} &  \bf{86.5} & \bf{81.3}\\
\bottomrule
\end{tabular}
\end{center}
\caption{Results on the RACE test set. BERT$_{\textsc{large}}$ and XLNet$_{\textsc{large}}$ results are from \newcite{yang2019xlnet}.}
\label{tab:roberta_race}
\end{table}

Results on the RACE test sets are presented in Table~\ref{tab:roberta_race}.
\ourmodel{} achieves state-of-the-art results on both middle-school and high-school settings.

\section{Related Work} \label{sec:relwork}

Pretraining methods have been designed with different training objectives, including language modeling~\cite{dai2015semi,peters2018deep,howard2018universal},  machine translation~\cite{mccann2017learned}, and masked language modeling~\cite{devlin2018bert,lample2019cross}. Many recent papers have used a basic recipe of  finetuning models for each end task~\cite{howard2018universal,radford2018gpt}, and pretraining with some variant of a masked language model objective.  However, newer methods have improved performance by multi-task fine tuning~\cite{dong2019unified}, incorporating entity embeddings~\cite{sun2019ernie}, span prediction~\cite{joshi2019spanbert}, and multiple variants of autoregressive pretraining~\cite{song2019mass,chan2019kermit,yang2019xlnet}.  Performance is also typically improved by training bigger models on more data~\cite{devlin2018bert,baevski2019cloze,yang2019xlnet,radford2019language}. Our goal was to replicate, simplify, and better tune the training of BERT, as a reference point for better understanding the relative performance of all of these methods.

\section{Conclusion} \label{sec:conclusion}

We carefully evaluate a number of design decisions when pretraining BERT models.
We find that performance can be substantially improved by training the model longer, with bigger batches over more data; removing the next sentence prediction objective; training on longer sequences; and dynamically changing the masking pattern applied to the training data.
Our improved pretraining procedure, which we call \ourmodel{}, achieves state-of-the-art results on GLUE, RACE and SQuAD, without multi-task finetuning for GLUE or additional data for SQuAD.
These results illustrate the importance of these previously overlooked design decisions and suggest that BERT's pretraining objective remains competitive with recently proposed alternatives.

We additionally use a novel dataset, \textsc{CC-News}, and release our models and code for pretraining and finetuning at: \url{https://github.com/pytorch/fairseq}.

\bibliography{main}
\bibliographystyle{acl_natbib}

\appendix

\begin{table*}[t]
\begin{center}
\begin{tabular}{lcccccccc}
\toprule
\bf  & \bf MNLI & \bf QNLI & \bf QQP & \bf RTE & \bf SST & \bf MRPC & \bf CoLA & \bf STS \\
\midrule 
\multicolumn{4}{l}{\ourmodelbase{}} \\
\quad + all data + 500k steps & 87.6 & 92.8 & 91.9 & 78.7 & 94.8 & 90.2 & 63.6 & 91.2 \\
\midrule
\multicolumn{4}{l}{\ourmodellarge{}} \\
\quad with \textsc{Books} + \textsc{Wiki} & 89.0 & 93.9 & 91.9 & 84.5 & 95.3 & 90.2 & 66.3 & 91.6 \\
\quad + additional data (\textsection\ref{sec:data}) & 89.3 & 94.0 & 92.0 & 82.7 & 95.6 & \textbf{91.4} & 66.1 & 92.2 \\
\quad + pretrain longer 300k & 90.0 & 94.5 & \textbf{92.2} & 83.3 & 96.1 & 91.1 & 67.4 & 92.3 \\
\quad + pretrain longer 500k & \textbf{90.2} & \textbf{94.7} & \textbf{92.2} & \textbf{86.6} & \textbf{96.4} & 90.9 & \textbf{68.0} & \textbf{92.4} \\
\bottomrule
\end{tabular}
\end{center}
\caption{
Development set results on GLUE tasks for various configurations of \ourmodel{}.
}
\label{tab:roberta_all_large_glue}
\end{table*}
\begin{table*}[t]
\begin{center}
\begin{tabular}{lccc}
\toprule
\bf Hyperparam  & \bf \ourmodellarge{} & \bf \ourmodelbase{} \\
\midrule 
Number of Layers & 24 & 12 \\
Hidden size & 1024 & 768 \\
FFN inner hidden size & 4096 & 3072 \\
Attention heads & 16 & 12 \\
Attention head size & 64 & 64 \\
Dropout & 0.1 & 0.1 \\
Attention Dropout & 0.1 & 0.1 \\
Warmup Steps & 30k & 24k \\
Peak Learning Rate & 4e-4 & 6e-4 \\
Batch Size & 8k & 8k\\
Weight Decay & 0.01 & 0.01 \\
Max Steps & 500k & 500k\\
Learning Rate Decay & Linear & Linear \\
Adam $\epsilon$ & 1e-6 & 1e-6 \\
Adam $\beta_1$ & 0.9 & 0.9 \\
Adam $\beta_2$ & 0.98 & 0.98 \\
Gradient Clipping & 0.0 & 0.0 \\
\bottomrule
\end{tabular}
\end{center}
\caption{
Hyperparameters for pretraining \ourmodellarge{} and \ourmodelbase{}.
}
\label{tab:pretraining_hyperparams}
\end{table*}
\begin{table*}[t]
\begin{center}
\begin{tabular}{lccc}
\toprule
\bf Hyperparam  & \bf RACE & \bf SQuAD & \bf GLUE \\
\midrule 
Learning Rate & 1e-5 & 1.5e-5 & \{1e-5, 2e-5, 3e-5\}\\
Batch Size & 16 & 48  & \{16, 32\}\\
Weight Decay & 0.1 & 0.01 & 0.1 \\
Max Epochs & 4 & 2 & 10 \\
Learning Rate Decay & Linear &Linear & Linear \\
Warmup ratio & 0.06 & 0.06 & 0.06 \\
\bottomrule
\end{tabular}
\end{center}
\caption{
Hyperparameters for finetuning \ourmodellarge{} on RACE, SQuAD and GLUE.
}
\label{tab:roberta_glue_finetune_hyperparams}
\end{table*}

\section*{Appendix for ``RoBERTa: A Robustly Optimized BERT Pretraining Approach"}

\section{Full results on GLUE}

In Table~\ref{tab:roberta_all_large_glue} we present the full set of development set results for \ourmodel{}.
We present results for a $\textsc{large}$ configuration that follows \bertlarge{}, as well as a $\textsc{base}$ configuration that follows \bertbase{}.

\section{Pretraining Hyperparameters}
Table~\ref{tab:pretraining_hyperparams} describes the hyperparameters for pretraining of \ourmodellarge{} and \ourmodelbase{} 

\section{Finetuning Hyperparameters}
\label{app:hyperparams}

Finetuning hyperparameters for RACE, SQuAD and GLUE are given in  Table~\ref{tab:roberta_glue_finetune_hyperparams}.
We select the best hyperparameter values based on the median of 5 random seeds for each task.

\end{document}